\documentclass[review]{elsarticle}

\usepackage[english]{babel}
\usepackage[T1]{fontenc}
\usepackage{etoolbox}
\makeatletter
\patchcmd{\ps@pprintTitle}{\footnotesize\itshape
      Preprint submitted to \ifx\@journal\@empty Elsevier
      \else\@journal\fi\hfill\today}{\scriptsize{Preprint submitted to Solar Energy \hfill \today}}{}{}

\def\ps@pprintTitle{%
   \let\@oddhead\@empty
   \let\@evenhead\@empty
   \let\@oddfoot\@empty
   \let\@evenfoot\@oddfoot
}
\makeatother
\usepackage{comment}

\usepackage{hyperref}
\usepackage{blindtext}
\usepackage{tcolorbox}
\usepackage{amsmath}

\usepackage{amsmath,amssymb,amsfonts}
\usepackage{algorithmic}
\usepackage[]{graphicx}
\graphicspath{{Figures/}}
\usepackage{lipsum}
\usepackage{balance}
\usepackage{textcomp}
\usepackage{xcolor}
\usepackage{multirow}
\usepackage{color}
\usepackage[document]{ragged2e}
\usepackage{lscape}
\usepackage{graphicx}

\usepackage{subcaption}
\usepackage{multirow}

\usepackage[switch]{lineno}

\modulolinenumbers[5]

\begin{document}

\begin{frontmatter}

\title{Analyzing the impact of feature selection on the accuracy of heart disease prediction}

\author[add1,add2]{Muhammad~Salman~Pathan}
\ead{salman.pathan@adaptcentre.ie}
\author[add3]{Avishek~Nag}
\ead{avishek.nag@ucd.ie}
\author[add4]{Muhammad~Mohisn~Pathan}
\ead{muhammd.mohsin@lumhs.edu.pk}
\author[add1,add2]{Soumyabrata~Dev\corref{mycorrespondingauthor}}
\cortext[mycorrespondingauthor]{Corresponding author. Tel.: + 353 1896 1797.}
\ead{soumyabrata.dev@ucd.ie}

\address[add1]{ADAPT SFI Research Centre, Dublin, Ireland}
\address[add2]{School of Computer Science, University College Dublin, Ireland}
\address[add3]{School of Electrical and Electronic Engineering, University College Dublin, Ireland}
\address[add4]{Institute of Biomedical Engineering and Technology, Liaquat University of Medical and Health Sciences, Pakistan}

\begin{abstract}
Heart Disease has become one of the most serious diseases that has a significant impact on human life. It has emerged as one of the leading causes of mortality among the people across the globe during the last decade. In order to prevent patients from further damage, an accurate diagnosis of heart disease on time is an essential factor. Recently we have seen the usage of non-invasive medical procedures, such as artificial intelligence-based techniques in the field of medical. Specially machine learning employs several algorithms and techniques that are widely used and are highly useful in accurately diagnosing the heart disease with less amount of time. However, the prediction of heart disease is not an easy task. The increasing size of medical datasets has made it a complicated task for practitioners to understand the complex feature relations and make disease predictions. Accordingly, the aim of this research is to identify the most important risk-factors from a highly dimensional dataset which helps in the accurate classification of heart disease with less complications. For a broader analysis, we have used two heart disease datasets with various medical features. Firstly, we performed the correlation and inter-dependence of different medical features in the context of heart disease. Secondly, we applied a filter-based feature selection technique on both datasets to select most relevant features (an optimal reduced feature subset) for detecting the heart disease. Finally, various machine learning classification models were investigated using complete and reduced features subset as inputs for experimentation analysis. The trained classifiers were evaluated based on Accuracy, Receiver Operating Characteristics (ROC) curve and F1-Score. The classification results of the models proved that there is a high impact of relevant features on the classification accuracy. Even with a reduced number of features, the performance of the classification models improved significantly with a reduced training time as compared with models trained on full feature set.
\end{abstract}


\end{frontmatter}

\section{Introduction}
\label{sec:1}
Heart disease is rapidly increasing across the globe. As per a research report published by the World Health Organization (WHO), in $2016$ approximately 17.90 million people died from heart
disease~\cite{nalluri2020chronic}. This much number accounts for approximately $30$\% of all deaths worldwide. 
 Nearly 55\% of the heart patient die during the first 3~years, and the treatment costs for heart disease are around 4\% of the annual healthcare expenditure.~\cite{manji2013cost}.  Observing the increasing stats, accurate
and timely detection and treatment of this serious illness is very essential for disease prevention and effective utilization of medical resources. 

Due to the recent technological advancements, the field of medical sciences has seen a remarkable improvement over time~\cite{saranya2019survey,sivapalan2022annet}. Specially, machine learning (ML) has been widely used in the field of
cardiovascular medicine and has established a potential space~\cite{haq2018hybrid}.
The basic framework of ML is built on models that take input data (such as text or images) and through the usage of some statistical analysis and mathematical optimizations provides the desired prediction results
(\textit{e.g.,} disease, no disease, neutral)~\cite{gavhane2018prediction}. 
   ML models can be trained on tons of raw electronic medical data gathered from low-cost wearable devices to allow efficient heart disease diagnosis with less resources and improved accuracy
~\cite{kumar2020analysis}.

During the training process, ML models require a large number of data samples to avoid overfitting~\cite{yeom2018privacy}. However, the inclusion of the large number of data features is not required for reasons related to
the curse of dimensionality~\cite{aremu2020machine,manandhar2018systematic}. Mostly, medical datasets cover related as well as redundant features. Unnecessary features do not contribute any meaningful information to the prediction task, and also creates noise in
the description of target (output class) which leads to prediction errors~\cite{pavithra2020review}. Furthermore, such features increase the complexity of ML models and make the system runs slowly due to increased training
time. To overcome the curse of dimensionality only those features which are closely related with the target should be selected/identified from datasets and provided as inputs to ML models~\cite{cai2018feature}. Relevant
feature selection can aid in performance improvement by decreasing the model complexity and increasing prediction accuracy which is very important in medical diagnosis~\cite{wang2021study}

Because of the benefits outlined previously, feature selection techniques are being actively used in the area of heart diseases and strokes~\cite{remeseiro2019review,pathan2020identifying,nwosu2019predicting}.

The contributions of this research are listed as follows:

\begin{itemize}

\item The study uses two datasets of heart disease patients from different sources to cover a broader study of medical features.
\item To perform the correlation and interdependence study between different features in datasets with respect to heart disease.
\item The identification of the most relevant medical features which aids in the prediction of heart disease using a filter-based feature selection technique.
\item Different ML classification models such as Logistic Regression (LR), Decision Tree (DT), Naive Bayes (NB), Random Forest (RF), Multi Layer Perceptron (MLP) \textit{etc.}, are used on the datasets to identify the suitable models for the problem.
\item The classification models were tested on full as well as the reduced feature subset to observe the impact of feature selection on the performance of models.
\item With the spirit of reproducible research, the code of this article is shared in GitHub.\footnote{\url{https://github.com/Sammyy092/analyzing-the-impact-of-feature-selection-on-heart-disease-prediction}.}
\end{itemize}

\section{Related Work}
\label{sec:rel_work}

ML has appeared to be an effective technique for assisting in the heart disease diagnosis, however the high dimensionality of datasets is a fundamental issue for ML prediction models. Feature selection is one of the techniques which is used to select only the most relevant features from datasets features that influence the disease outcome most. The identification of the most important features from the high dimensional datasets is an important aspect that can improve the accuracy of prediction models hence reduce the number of medical injuries.

In~\cite{zhang2019stroke}, Zhang \textit{et al.} developed an efficient feature selection technique called weighting-and ranking-based hybrid feature selection (WRHFS) to determine the risk of heart stroke. For the weighing and
ranking of features, WHRFS used a variety of filter-based feature selection techniques such as fisher score, information gain and standard deviation. The proposed technique selected $9$ important input features out
of $28$ based on the knowledge provided for heart stroke prediction. In another research~\cite{le2018automatic}, the authors worked on the extraction of relevant risk factors form a large feature space for an
efficient heart disease prediction. The features were selected based on their individual ranks. The authors used Latent Feature Selection (ILFS) method to rank the features which is a probabilistic latent graph-based
feature selection technique. The results of the model were competitive using only half of the features from the set of $50$. In~\cite{zhang2018risk}, a feature selection model for detecting the risk of heart disease
is proposed. The proposed model combined the glow-worm swarm optimization algorithm based on the standard deviation of the features to extract the quality features from a electronic healthcare record (EHR) of a community
hospital in Beijing. $6$ features including high blood pressure, Alkaline Phosphatase (ALP), age and Lactate Dehydrogenase (LDH) were indicated as important features to detect stroke excluding the family hereditary
factors. The authors of~\cite{al2021identifying} focused on finding the most relevant features from EHR to predict the early-stage risk of death from heart disease. The authors used minimum redundancy maximum (mRmR)
relevance and recursive feature elimination (RFE) feature selection approaches based on NB  for the selection of features. Two medical features \textit{i.e.,} Serum Creatinine and Ejection Fraction were ranked higher by
both feature selection technique as compared to other. When provided to a prediction model as input, the selected features proved out to be most important as an overall accuracy of $80$\% was achieved. Singh \textit{et al.} ~\cite{singh2017stroke}, proposed an efficient approach for stroke prediction using the Cardiovascular Health Study (CHS) dataset. They used DT algorithm for feature selection and then principal component analysis
(PCA) technique for reducing the dimensionality of feature space. Finally, the MLP network was used to construct the classification model. The model trained on the optimal feature set achieved 97.7\% accuracy in detecting
the stroke and outperformed other techniques in comparison. A wrapper based Genetic Algorithm (GA) is used in~\cite{gokulnath2019optimized} to select the most significant features to detect heart disease. The proposed
feature selection algorithm identifies $7$ features out of $16$ to detect heart disease from Cleveland heart disease dataset. The resultant features were supplied to support vector machine (SVM) for the
accuracy evaluation. The classifier acquired 88.34\% using the reduced feature set whereas only 83.34\% was achieved when using whole dataset features. In terms of ROC curve, the GA-SVM performed well also when compared
with the various existing feature selection algorithms also. This study~\cite{zhang2021heart} proposes a new heart disease prediction model by combining ML with deep learning techniques. The least absolute shrinkage and
selection operator (LASSO) penalty method based on LinearSVC was applied as the feature selection module to generate a feature subset closely related to target. $12$ most relevant features were chosen from dataset
obtained from Kaggle and inputted to the MLP network. As per the experimental results, the proposed model obtained an accuracy of 98.56\% with 99.35\% recall and 97.84\% precision. In~\cite{haq2018hybrid}, a ML based heart
disease diagnosis system is proposed. Seven popular classifiers LR, k-Nearest Neighbor (K-NN), MLP, SVM, NB, DT, and RF were used for the classification of heart disease patients. Three feature selection algorithms RelieF,
mRMR, and LASSO were used to select highly correlated features with target class. It was observed that the classification performance of models increased in terms of accuracy and computation time using the feature
selection techniques. The LR model showed best accuracy of $89$\% when used with RelieF. The main objective of this research~\cite{hasan2021comparing} was to predict the heart disease using minimal subset of
features and adequate accuracy. To achieve this objective, the authors employed a two-stage feature subset retrieving technique. Three popular feature selection techniques \textit{i.e.,} (embedded, filter, wrapper) were
used to extract a feature subset based on a boolean process-based common ``True'' condition. To select the suitable prediction model, RF, SVM, K-NN, NB, XGBoost and MLP models were trained on the data. The experimental
results showed that XGBoost classifier integrated with wrapper technique provided the best prediction results for heart disease. A comparative analysis of different classifiers was performed in~\cite{reddy2021heart} for
the classification of the heart disease with minimal attributes. ML classifiers such as NB, LR, sequential minimal optimization (SMO), RF \textit{etc.,} were trained for the accurate detection of heart disease. To obtain
the optimal feature subset, RelieF, chi-squared and correlation-based feature subset evaluator were utilized. $10$ features were selected from the set of $13$ to train the classifiers. The SMO classifier
achieved the highest accuracy of 86.468\% when inputted with the optimal feature set obtained by chi-squared feature selection technique. 

Despite their relevance, one major drawback of existing works on heart disease prediction is the lack of systematic guidance when selecting the input features for the development of prediction models which is an important aspect in terms of predictive performance. Previous research proposals chose features mostly in an impromptu manner without incorporating latest medical research findings. Mostly the focus is on the prediction models and their final prediction performance. However, a very less attention is paid on the correlation between different medical features and their individual importance in the prediction of heart disease. A few works present analysis of medical features but for the purpose of heart disease detection only. This research aims at addressing the ineffective feature selection in previous studies on heart disease prediction. Two heart disease patient datasets collected from different sources were utilized in this research to cover a broader study of features related to heart disease and to identify various medical procedures. To further analyze the role of each parameter in the prediction task, we obtain the interdependence and importance of the collected set of medical features. A detailed analysis of ML models trained on both full and selected feature set is provided to analyze the impact of feature selection techniques on the prediction performance as well as the identification of suitable classifiers for the specified problem.

\section {Proposed Methodology}
\begin{figure}[htb]
    \centering
    \includegraphics[width=0.30\textwidth]{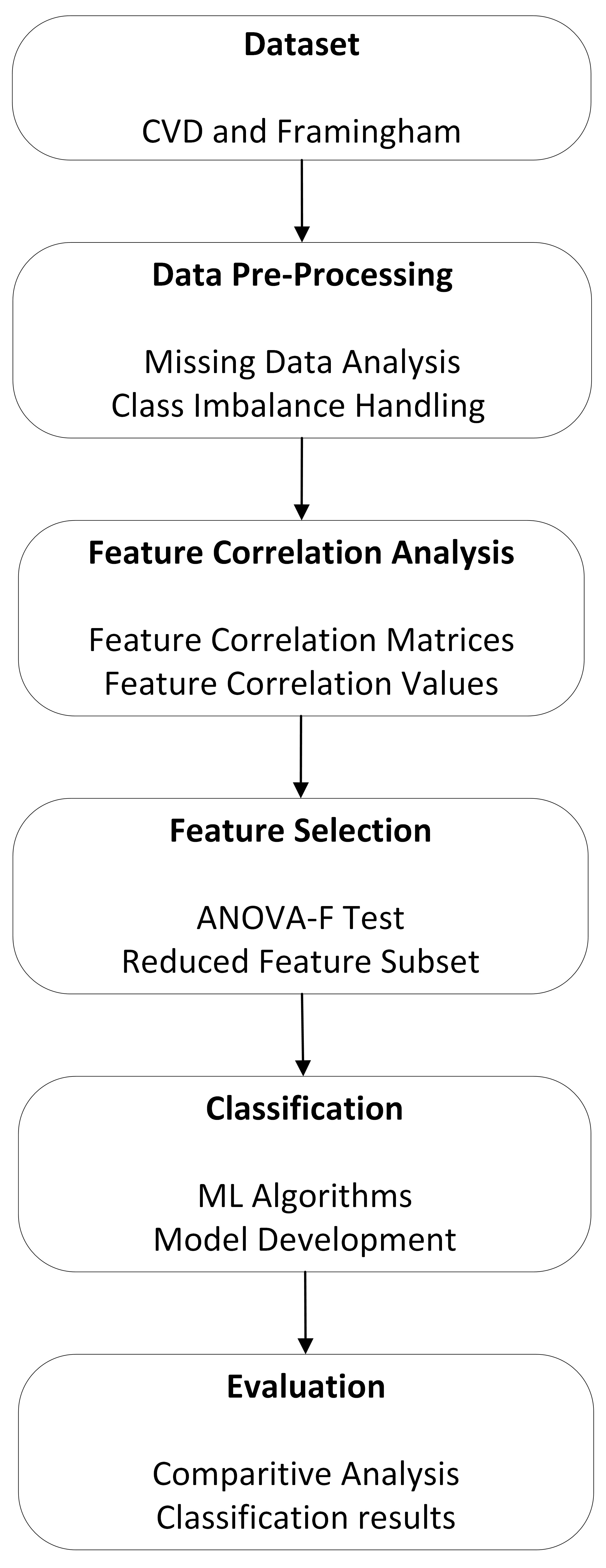}
    \caption{Flowchart of the proposed methodology describing each step for heart disease prediction.}
    \label{fig:flowchart}
\end{figure}

This research paper highlights the importance feature selection in the accurate classification of heart disease.  Figure~\ref{fig:flowchart} demonstrates the workflow of the proposed methodology for heart disease prediction.

\subsection{Datasets}
\label{sec:data}

In this research, two datasets named as cardiovascular disease (CVD) and Framingham were utilized to study the impact of different features on the occurrence of heart disease and to develop ML-based system for heart disease detection. The study uses two datasets to cover a broader study of medical features and various clinical pathways used for the detection of heart stroke. The datasets were collected from different sources. The datasets contained some main medical features like `age', `hypertension', 'glucose levels', 'blood pressure', `cholesterol' \textit{etc.} which are closely related to the occurrence of disease and provides a great flexibility for heart disease analysis. The datasets were chosen based on two criteria. The first criterion was the variance in the medical procedures, so to study the different medical procedures and the role of each feature in the context of heart disease. Secondly, the datasets were chosen based on the data availability. Datasets from different sources possess different amount of data and collection of features. So, we have chosen datasets which were offering a good volume of data and having a level of similarity in terms of features.

\subsubsection{CVD}
The CVD dataset is controlled by McKinsey \& Company which was a part of their healthcare hackathon\footnote{https://datahack.analyticsvidhya.com/contest/mckinsey-analyticsonline-hackathon/}. The dataset can be accessible from a free dataset repository\footnote{https://inclass.kaggle.com/asaumya/healthcare-dataset-stroke-data}. The collected dataset included $29072$ patient observation with $12$ data features. $11$ of them are the common clinical symptoms and are considered as input features whereas the $12$\textit{th} feature `stroke' is the target feature indicating whether a patient has had stroke or not. 
The complete description of data features for CVD dataset is given in Table~\ref{tab:cvd_feat}.

\subsubsection{Framingham}
The Framingham dataset was created during an ongoing cardiovascular study involving the residents of Framingham, Massachusetts, and is available at the Kaggle website\footnote{https://www.kaggle.com/amanajmera1/framingham-heart-study-dataset}. The dataset is mostly used in classification tasks to identify whether a patient has a chance to develop coronary heart disease (CHD) in $10$ years. The dataset contains $4,240$ patient records and $15$ features, where each feature indicates a risk factor. $14$ input features were used to detect the decisional feature \textit{i.e.,} 10-year risk of CHD. Table~\ref{tab:fram_feat} shows the description about the data features in Framingham dataset.
\begin{table}[]
\centering
\begin{tabular}{|c|c|}
\hline
\textbf{Attribute}           & \textbf{Description}                                                                                                                            \\ \hline
i.d              &  patient's i.d                                                            \\ \hline
gender              & includes  ("male": 0, "female": 1, "other": 2)                                                                                         \\ \hline
age                 & patient's age (continuous)                                                                                                                          \\ \hline
hypertension        & suffering from hypertension  ("yes":1, "no":0)                                                                                         \\ \hline
heart \_disease     & suffering heart disease ("yes":1, "no":0)                                                                                              \\ \hline
ever\_married       & marital status of patient ("yes":1,  "no":0)                                                                                           \\ \hline
work\_type          & \begin{tabular}[c]{@{}c@{}}job status ("children":0, "govt\_job":1,\\ "never\_worked":2, "private":3,\\ "self\_employed":4)\end{tabular} \\ \hline
residence\_type     & ("rural:0, "urban":1)                                                                                                                  \\ \hline
avg\_glucose\_level & average glucose level of blood (continuous)                                                                                                         \\ \hline
bmi                 & body mass index (decimal value)                                                                                                        \\ \hline
smoking\_status     & ("never smoked":0, "formerly smoked":1, "smokes":2)                                                                                    \\ \hline
stroke              & ("yes":1, "no":0)                                                                                                                      \\ \hline
\end{tabular}
\caption{Description of features CVD dataset}
\label{tab:cvd_feat}
\end{table}

\begin{table}[]
\centering
\begin{tabular}{|c|c|}
\hline
\textbf{Attribute}       & \textbf{Description}                                                                                           \\ \hline
age             & patient's age (continuous)                                                                            \\ \hline
male            & ("male":0, "feamale":1)                                                                               \\ \hline
education       & level of education (1 to 4)                                                                           \\ \hline
currentSmoker  & ("smoker":1, "non smoke":0)                                                                           \\ \hline
CigsPerDay      & \begin{tabular}[c]{@{}c@{}}average number of ciggerates consumed \\ per day (continuous)\end{tabular} \\ \hline
BPMeds          & on blood pressure medication ("yes":1, "no":0)                                                        \\ \hline
prevalentStroke & previous stroke history ("yes": 1, "no":0)                                                            \\ \hline
prevalenHyp     & hypertensive ("yes":1, "no":0)                                                                        \\ \hline
diabetes        & previous diabetes history("yes":1, "no":0)                                                            \\ \hline
totCHol       & cholestrol level (continuous)                                                                         \\ \hline
sysBP           & systolic blood pressure (decimal)                                                                     \\ \hline
diaBP          & diastolic blood pressure (decimal)                                                                    \\ \hline
BMI             & body mass index (decimal)                                                                             \\ \hline
HeartRate      & heart rate measure (continuous)                                                                       \\ \hline
glucose         & glucose level (continuous)                                                                            \\ \hline
TenYearCHD  & target ("yes": 1, "no": 0)                                                                            \\ \hline
\end{tabular}
\caption{Description of features Framingham dataset}
\label{tab:fram_feat}
\end{table}

\subsection{Pre-Processing}
\label{sec:preprocess}

Data pre-processing is one of the important part of ML life cycle as it makes data analysis easy and increases the accuracy and speed of the ML algorithms~\cite{huang2015empirical} . We applied some pre-processing steps as
the collected dataset were having missing values and class imbalance problems. Referring the CVD dataset, the dataset contained a total of $43400$ patient records out of which $14754$ values were missing or null.
Whereas, $4240$ patient records were available for framingham dataset of which $645$ values were null. A null value does not necessarily mean that the value does not exist, but it is unknown. In medical
datasets, mostly the null or missing value is usually due to a lack of collection or the practitioner may not consider the observation since the medical test is considered to be low yield for the patient. Data imputation
methods are useful in handling the missing data, however their usage in medical field is limited and specific efficacy for disease detection is not clear~\cite{sachan2021evidential}. Most of the times, researchers do not
consider the observations with missing values and drop the incomplete cases intentionally, since the traditional data imputation methods are not sufficient to capture the missing data complexities in health care
applications~\cite{wang2022application,stavseth2019handling}. However, only a deep knowledge of specific disease will likely aid in the selection of the suitable data imputation methods. As per the mentioned analysis, we
dropped all the observations with null value from both the datasets to avoid any accuracy biases.

Furthermore, looking at the class distribution, both datasets were highly unbalanced in nature. Only $548$ patients out of 29,072 in CVD dataset had stroke conditions, whereas 28,524 patients had no occurrence of
stroke. In framingham dataset, only $557$ patient records showed the risk of CHD out of $3101$. The unbalanced nature of the datasets leads to classification errors during the training of ML
models~\cite{buda2018systematic}. As a result, we adopted a 'Random Down-Sampling' technique to mitigate the adverse effects caused by unbalanced data. We made two classes referred as `minority' and `majority' classes. The
patients with heart disease were included in minority class, whereas the patients having no symptoms were included in majority class. In the case of CVD dataset, $548$ observations were included into the minority
class and the remaining 28,524 were considered as majority class. We created a balanced dataset of $1096$ observations by selecting all $548$ observations from minority class and $548$ random observations
from a total of 28,524 majority cases. Same process was performed for framingham dataset where $557$ random observations from $3101$ majority cases were derived making a total of $1114$ observations in a
balanced dataset shape. In this way, two balanced datasets were made to study the features importance and disease classification in an efficient manner.

\subsection{Feature Correlation Analysis}
\begin{figure}[htb]
\centering
\subfloat[\centering CVD Dataset]{
\includegraphics[height=0.44\textwidth]{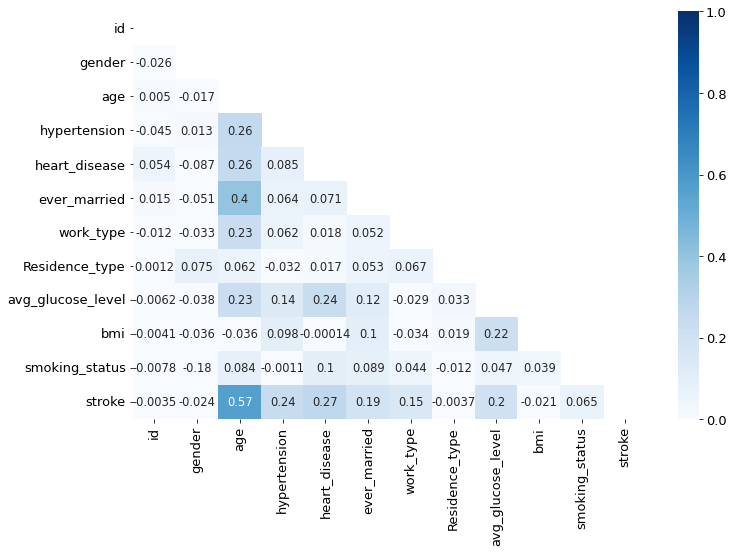}}
\subfloat[\centering Framingham Dataset]{
\includegraphics[height=0.44\textwidth]{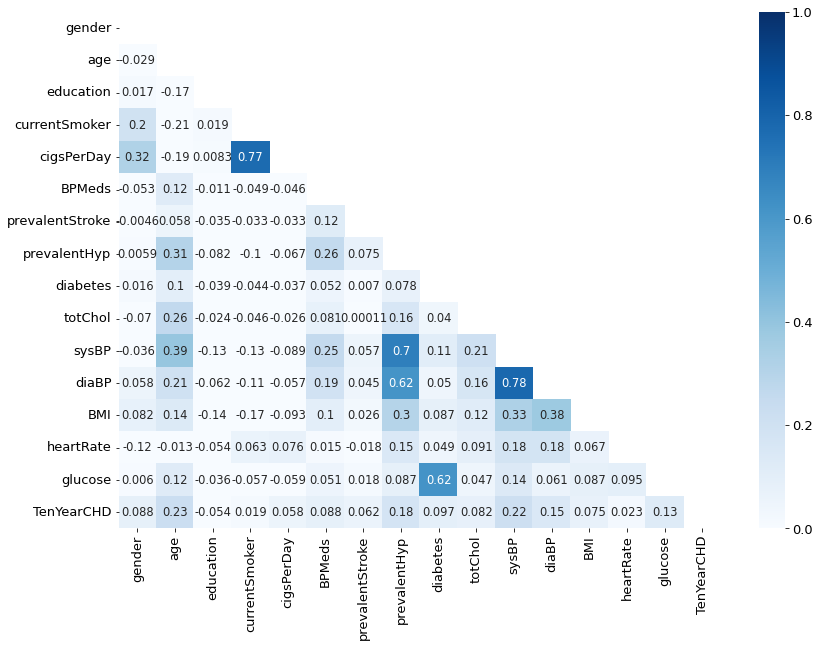}}
\caption{The correlation values for each medical features and the target heart disease for both datasets. 
}
\label{fig:corrmatrix}
\end{figure}

Feature correlation is a method which helps in understanding the underlying relationships between various data features present in a dataset. Feature correlation can be useful in many ways such as determining the inter-dependencies between the data features and how each feature effects the output feature \cite{gopika2018correlation}. We obtained the correlation values between the data features by calculating the correlation coefficients of the feature matrix $M$ having dimension $p \times q$, denoted as: $M=[v_1, v_2,\ldots, v_q]$, where $v_1, v_2,\ldots, v_q$ are the vectors having $q$ number of features. $p$ indicates the length of the vector, where each vector is a complete medical procedure at a specific time. The computed correlation values between different medical features and the target disease for each dataset are shown Figure~\ref{fig:corrmatrix}.

As we can see from the Figure~\ref{fig:corrmatrix}, of the $11$ features of CVD dataset $4$ features are having a positive correlation with the decision feature \textit{i.e.}, ‘stroke’. Features ‘age’, ‘hypertension’, 'heart\_disease' and 'avg\_glucose\_lvl' are having values $0.57$, $0.24$, $0.27$ and $0.2$ when correlated with 'stroke' showing a significant correlation. Similarly, for framingham dataset, features 'age', 'sysBP', 'prevalentHyp', 'diabBP' and 'glucose' showed positive values of $0.23$, $0.22$, $0.18$, $0.15$ and reflect the motif of the desired output feature 'TenYearCHD'. For both the datasets, features like 'gender', 'bmi', 'heart rate' and other non-medical features like smoking habits, education, social status and living standards showed very less correlation with the output feature having no or very less effect on the output. Overall, the common medical features like 'age', 'hypertension' and 'glucose' in both datasets are closely related with the outcome and can be considered as the important risk factors.

As per medical research findings, with aging, major changes can be observed in the heart and blood vessels. For example, the heartbeat rate is not as fast during any physical activity as it could when you are younger. The age-related changes may raise a person's risk of heart disease according to National Heart, Lung, and Blood Institute Trusted Source \cite{williams2006report}. Hypertension is an established risk factor for stroke, ischemic heart disease and renal dysfunction \cite{escobar2002hypertension}. Hypertension causes the blood pressure over the normal range. The higher blood pressure levels make the arteries less elastic and decreases the oxygen and blood flow towards the heart which potentially leads to a heart disease. The diabetic patients are more likely to develop heart disease at an earlier stage. High blood glucose from diabetes causes stronger contraction of blood vessels that control your heart and blood vessels which leads to heart disease \cite{huxley2006excess}. Over time, this process can lead to a heart stroke.
\label{sec:corranalysis}

\subsection{Feature Selection}

The main motivation of this research is to select the medical features that can improve the accuracy of heart disease prediction. Feature selection is the process of selecting a subset of most relevant features from a larger collection of original features, that influence the outcome most. The advantages of feature selection includes: data quality improvement, less computational time by prediction model, predictive performance improvement and efficient data collection process.

In this work we have used a filter-based feature selection technique namely, ANOVA-F test to identify most important features from both datasets. Filter-based feature selection techniques employ the use of statistical
methods such as similarity, dependence, information, distance to point out the important dependencies or correlation between the input and the target features~\cite{bommert2020benchmark}. Analysis of Variance
(ANOVA) is a collection of parametric statistical models and their estimation procedures that determines if the means of two or more samples of data originate from the same distribution. F-test also known as F-statistic,
is a set of statistical tests that uses some statistical techniques to calculate the ratio of variance values such variance of two separate samples \textit{etc.} The ANOVA method is a type of F-statistic referred here as
an ANOVA f-test. It is a univariate statistical test where each feature is compared to the target feature, to see whether there is any statistically significant relationship between them~\cite{mishra2019application}.
Mostly, ANOVA is used in such classification tasks where the type of input features is numerical the target feature is categorical.

The ANOVA-F test can be implemented in python language using the f\_classif() function provided by scikit-learn library. The f\_classif() function is used in selecting the most important features (features with largest values) via the SelectKBest class. SelectKBest is a method made available in the scikit-learn which takes a scoring function and ranks the features by these scores. Here The scoring function is f\_classif() \textit{i.e.,} ANOVA-F test and we have defined SelectKBest class to identify most important features from datasets. The equation to obtain ANOVA-F values is given below:

\[variance\_between\_groups=\frac{\sum_{i=1}^{j}j_{i}(\bar{K}_{i}-\bar{K})^2}{(S-1))}\]

\[variance\_within\_groups=\frac{\sum_{i=1}^{S}\sum_{p=1}^{j_{i}}({K}_{ip}-\bar{K}_{i})^2}{(N-S))}\]

\[F\_value= \frac{variance\_between\_groups}{variance\_within\_groups}\]

where $N$ is the overall sample size, $S$ is the number of groups, $j_{i}$ is the number of observations in the $jth$ group, $\bar{K}_{i}$ is the $ith$ group sample mean, $\bar{K}$ is the overall mean of the data,  $K_{ip}$ is the $pth$ observation in the $ith$ out of $S$ groups

\begin{figure}[htb]
\centering
\subfloat[\centering CVD Dataset]{
\includegraphics[height=0.22\textwidth]{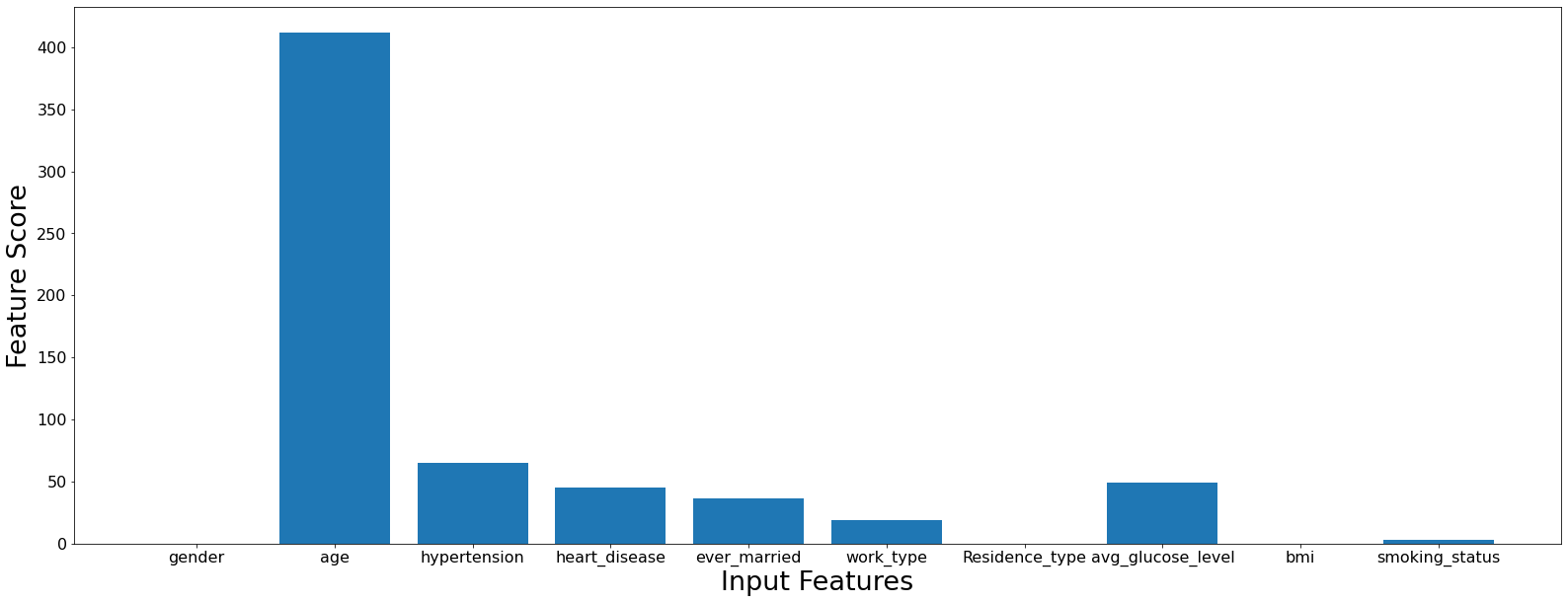}}
\subfloat[\centering Framingham Dataset]{
\includegraphics[height=0.22\textwidth]{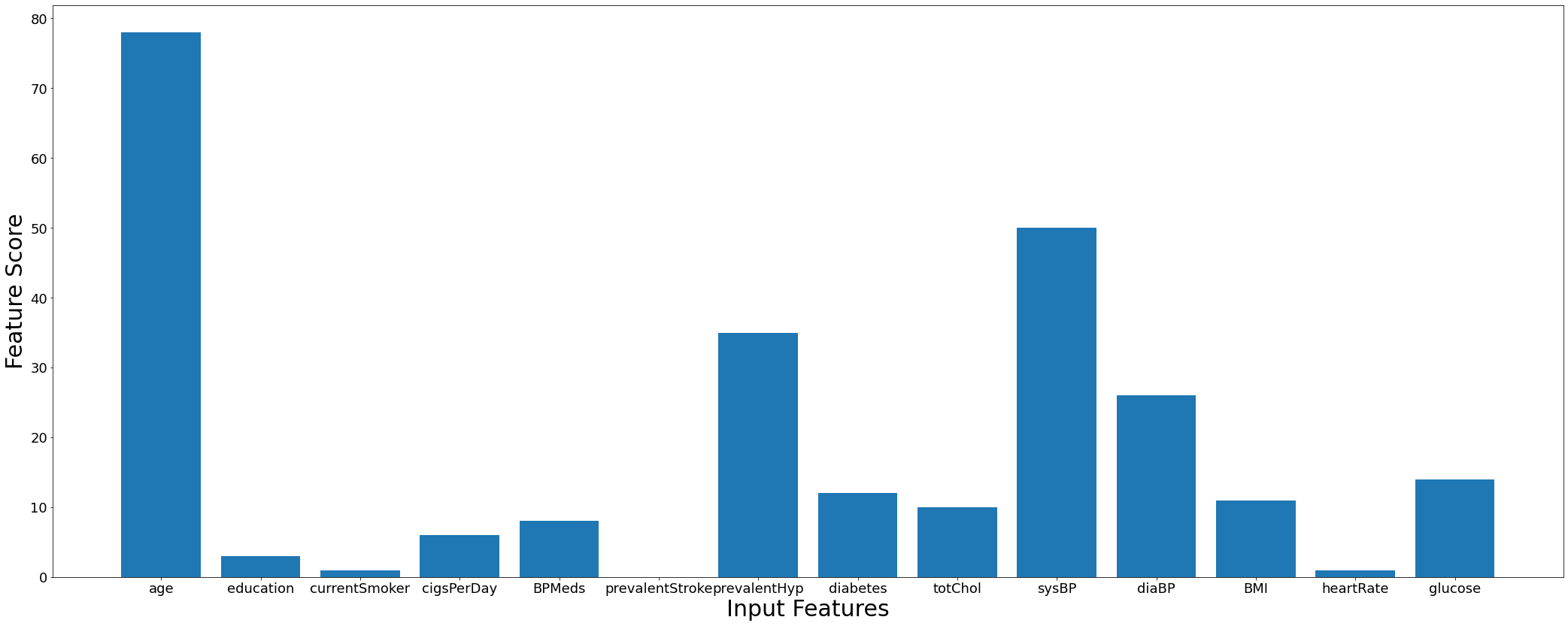}}
\caption{Feature importance scores for each feature in both datasets. 
}
\label{fig:featscore}
\end{figure}

The feature importance scores obtained using the ANOVA-F test are shown in figure~\ref{fig:featscore} (a) and (b) for both datasets. According to the statistics in figure~\ref{fig:featscore} (a), the most important features for predicting 'stroke' are 'age', 'hypertension', 'heart\_disease' and 'avg\_glucose\_lvl' possessing suitable scores when related with the outcome. However, features `gender', `bmi', 'residence\_type' and 'smoking\_status' showed less or 0 significance for the feature `stroke'.  Looking at ~\ref{fig:featscore} (b), we can observe that features `age', `prevalentHyp', `diabetes', `sysBP', `diaBP' and `glucose' obtains highest scores as compared to the other features of the dataset when related to `TenYearCHD'. Looking at the importance values of the features for each dataset, we can observe a similarity with the correlation results listed in section~\ref{sec:corranalysis} \textit{i.e.}, in most cases the features related with age, hypertension, glucose, blood pressure has a significant influence in the prediction of the heart disease. Similarly, the features identified using ANOVA-F test are also listed as the potential risk factors for heart disease as cited by the American Heart Association \cite{benjamin2019heart}. 

\section{Evaluation Matrices}
We have used three popular performance evaluation metrices \textit{i.e.,} Accuracy, F1-score and ROC to evaluate the performance of ML classification models~\cite{dev2017nighttime}. Confusion matrix is a table that helps ML practitioners to describe the performance of a classification model. Confusion matrix consists of four used to determine the performance matrices of a classifier and can be described as (1) True Positive ($TP$) test result that correctly classify the presence of heart disease in patient, (2) True Negative ($TN$) test result that correctly classify the absence of heart disease in patient, (3) False Negative ($FN$) test result that wrongly classify that a particular patient does not have heart disease and (4) False Positive ($FP$) test result which wrongly classify that a particular patient has heart disease. In medical field, $FN$ are considered as most harmful predictions. For a given dataset of size $n$, accuracy is measured as

\[Accuracy=(TP+TN)/(TP+FP+FN+TN)
\]
F1-Score is the harmonic mean of Precision and Recall.
\[Precision = TP/(TP + FP)
\]
\[Recall = TP/(TP + FN)
\]
\[F1-Score = 2(Precision\times Recall)/(Precision\times Recall)
\]

The Receiver Optimistic Curves (ROC) examine the classification capability of a classification model. It evaluates the “true positive rate” and “false positive rate” in a ML model output.
\[TPR = T P /(T P + F N)
\]
\[FPR = F P /(F P + T N)
\]

\section{Results and Discussions}
In this section, we will discuss the performance of the selected classification models from different perspectives. First, we checked the performance of model individually for both datasets with full features to examine which models work well for each dataset. Secondly, we evaluated the performance of the models on the selected set of feature to analyze the effect of feature selection technique on the accuracy of the classifiers. The classifiers performance was checked using the Accuracy, F1-score and ROC evaluation matrices.

\subsection{Classification results using full feature set}
In this section, all the ML models were tested on both datasets using full set of features to predict the binary disease outcome.
We trained all the prediction models on entire data with 80\% training and 20\% testing subsets. The overall computational time consumed during the training of prediction models was 10.98 iterations per second (it/s) for CVD dataset and 24.20 iterations per second (it/s) using framingham dataset. Table~\ref{tab:CVD_fulfeat} and ~\ref{tab:Fram_fulfeat} shows the binary classification results of the  ML model in predicting the heart disease for both datasets.

\begin{table}[]
\centering
\begin{tabular}{l|c|c|c|c}
\hline
\textbf{Model}                           & \textbf{Accuracy} & \textbf{Balanced Accuracy} & \textbf{ROC AUC} & \textbf{F1-Score} \\ \hline
Perceptron                      & 0.73     & 0.74              & 0.74    & 0.73     \\ \hline
SGD Classifier                  & 0.72     & 0.73              & 0.73    & 0.71     \\ \hline
Logistic Regression             & 0.73     & 0.73              & 0.73    & 0.73     \\ \hline
Quadratic Discriminant Analysis & 0.72     & 0.73              & 0.73    & 0.72     \\ \hline
Linear SVC                      & 0.72     & 0.72              & 0.72    & 0.72     \\ \hline
SVC                             & 0.71     & 0.72              & 0.72    & 0.71     \\ \hline
Nu SVC                          & 0.71     & 0.72              & 0.72    & 0.71     \\ \hline
Nearest Centroid                & 0.71     & 0.72              & 0.72    & 0.71     \\ \hline
Calibrated Classifier CV        & 0.71     & 0.72              & 0.72    & 0.71     \\ \hline
Bernoulli NB                    & 0.71     & 0.72              & 0.72    & 0.71     \\ \hline
Gaussian NB                     & 0.71     & 0.71              & 0.71    & 0.71     \\ \hline
Passive Aggressive Classifier    & 0.71     & 0.71              & 0.71    & 0.71     \\ \hline
Ridge Classifier CV             & 0.70     & 0.71              & 0.71    & 0.70     \\ \hline
Ridge Classifier                & 0.70     & 0.71              & 0.71    & 0.70     \\ \hline
Linear Discriminant Analysis    & 0.70     & 0.71              & 0.71    & 0.70     \\ \hline
Random Forest Classifier        & 0.70     & 0.70              & 0.70    & 0.70     \\ \hline
AdaBoost Classifier             & 0.70     & 0,70              & 0.70    & 0.70     \\ \hline
KNeighbors Classifier           & 0.69     & 0.69              & 0.69    & 0.69     \\ \hline
Bagging Classifier              & 0.68     & 0.68              & 0.68    & 0.68     \\ \hline
Extra Tree Classifier           & 0.66     & 0.66              & 0.66    & 0.66     \\ \hline
LGBM Classifier                 & 0.65     & 0.66              & 0.66    & 0.65     \\ \hline
XGB Classifier                  & 0.65     & 0.65              & 0.65    & 0.65     \\ \hline
Decision Tree Classifier        & 0.61     & 0.61              & 0.61    & 0.61     \\ \hline
Label Spreading                 & 0.60     & 0.60              & 0.60    & 0.60     \\ \hline
Label Propagation               & 0.60     & 0.59              & 0.59    & 0.60     \\ \hline
Dummy Classifier                & 0.46     & 0.46              & 0.46    & 0.46     \\ \hline
\end{tabular}
\caption{Classification results of various ML models for CVD dataset using full feature set}
\label{tab:CVD_fulfeat}
\end{table}

\begin{table}[]
\centering
\begin{tabular}{l|c|c|c|c}
\hline
\textbf{Model}                           & \textbf{Accuracy} & \textbf{Balanced Accuracy} & \textbf{ROC AUC} & \textbf{F1-Score} \\ \hline
Linear SVC                      & 0.66     & 0.67              & 0.67    & 0.66     \\ \hline
Linear Discriminant Analysis    & 0.66     & 0.67              & 0.67    & 0.66     \\ \hline
Calibrated Classifier CV        & 0.66     & 0.67              & 0.67    & 0.66     \\ \hline
Ridge Classifier CV             & 0.66     & 0.67              & 0.67    & 0.66     \\ \hline
Ridge Classifier                & 0.66     & 0.67              & 0.67    & 0.66     \\ \hline
Logistic Regression             & 0.65     & 0.66              & 0.66    & 0.65     \\ \hline
Nearest Centroid                & 0.64     & 0.66              & 0.66    & 0.64     \\ \hline
KNeighbors Classifier           & 0.64     & 0.65              & 0.65    & 0.64     \\ \hline
Random Forest Classifier        & 0.64     & 0.65              & 0.65    & 0.64     \\ \hline
Bernoulli NB                    & 0.63     & 0.64              & 0.64    & 0.63     \\ \hline
LGBM Classifier                 & 0.63     & 0.64              & 0.64    & 0.63     \\ \hline
AdaBoost Classifier             & 0.63     & 0.64              & 0.64    & 0.63     \\ \hline
Extra Tree Classifier           & 0.62     & 0.63              & 0.63    & 0.62     \\ \hline
XGB Classifier                  & 0.62     & 0.63              & 0.63    & 0.61     \\ \hline
Decision Tree Classifier        & 0.61     & 0.62              & 0.62    & 0.61     \\ \hline
Bagging Classifier              & 0.60     & 0.61              & 0.61    & 0.59     \\ \hline
SGD Classifier                  & 0.60     & 0.60              & 0.60    & 0.60     \\ \hline
Gaussian NB                     & 0.57     & 0.60              & 0.60    & 0.53     \\ \hline
Passive Aggressive Classifier    & 0.58     & 0.59              & 0.59    & 0.57     \\ \hline
Nu SVC                          & 0.59     & 0.59              & 0.59    & 0.59     \\ \hline
Extra Tree Classifier           & 0.59     & 0.59              & 0.59    & 0.59     \\ \hline
SVC                             & 0.58     & 0.58              & 0.58    & 0.58     \\ \hline
Quadratic Discriminant Analysis & 0.55     & 0.58              & 0.58    & 0.51     \\ \hline
Perceptron                      & 0.57     & 0.55              & 0.55    & 0.55     \\ \hline
Label Propagation               & 0.54     & 0.54              & 0.54    & 0.53     \\ \hline
Label Spreading                 & 0.53     & 0.53              & 0.53    & 0.53     \\ \hline
Dummy Classifier                & 0.52     & 0.52              & 0.52    & 0.52     \\ \hline
\end{tabular}
\caption{Classification results of various ML models for Framingham dataset using full feature set}
\label{tab:Fram_fulfeat}
\end{table}

Looking at the classification results listed In Table~\ref{tab:CVD_fulfeat}, the highest accuracy reported was $0.73$ achieved by MLP for CVD dataset with ROC of $0.74$ and F1-Score of $0.73$. Along with MLP other classifiers like LR, SVC and RF worked well and provided reasonable prediction accuracy with full feature set. The reason behind the improved accuracy achieved by MLP is that it is good at discovering patterns from complex medical datasets. Furthermore, this network model is good at generalizing data without having the prior domain knowledge. The worst results were obtained by the dummy classifier \textit{i.e.}, only a $0.46$ accuracy when predicting heart stroke. Possible reasons behind poor classification result is that the dummy classifier makes predictions using simple rules which is not useful when dealing with real world problems. The classification results with the same techniques for the framingham dataset are shown in Table~\ref{tab:Fram_fulfeat}. The accuracy results were not very good as the highest accuracy achieved was 0.66 with 0.67 ROC and 0.66 F1-score. Other algorithms like Linear Discriminant Analysis (LDA), LR and ridge
classifier performed the similar. The reason behind weak results might be the range of values between the data features. Feature scaling helps in normalizing the data within a particular range, which can improve the
results of the models in general~\cite{thara2019auto,jain2021validating}. However, any data manipulation strategy in medical studies may introduce significant biases, that is why we have kept all the feature values unchanged.

\subsection{Classification results using reduced feature set}
Given the goal of identifying the potential bio-markers and to analyze the impact of feature selection technique on the classification accuracy, we selected the most prominent features from the full feature space based on individual feature scores. The features impacting the outcome most for each dataset were identified by ANOVA-F test as shown in Figure~\ref{fig:featscore} (a) and (b). As per the feature scores, $4$ features \textit{i.e.,} \{age, hypertension, heart\_disease, avg\_glucose\_lvl\} were selected for CVD dataset out of $11$. Only $5$ features out of $15$ from framingham dataset \textit{i.e.,} \{age, prevalentHyp, sysBp, diaBp, glucose\} were chosen considering the feature weights obtained using ANOVA-F test. We evaluated the performance of each classification model using only the selected features as inputs. Table~\ref{tab:CVD_redufeat} shows the classification performance of each model using the reduced feature subset from CVD dataset. The analysis showed that even after limiting the number of features, ML models showed better performance as compared to the models using full feature set. The highest accuracy achieved was $0.74$ by SVC model with $0.74$ F1-Score and $0.74$ ROC with only $4$ input features. Considering Table~\ref{tab:Fram_redufeat} results, the highest accuracy achieved is $0.71$, which is higher than all the accuracy results using full feature set for framingham dataset. Furthermore, the models trained on reduced feature set also consumed less computational time \textit{i.e} only 3.86 iterations per second(it/s) using CVD and 15.52 iterations per second(it/s) using framingham dataset. We have also validated our findings by comparing our work with other published proposals \cite{dev2022predictive, beunza2019comparison} where same datasets were used with full feature set and the obtained accuracy results were less or equal to the results that we obtained using reduced feature set. Overall, the experimental results proved that the performance of the ML models increased significantly by using only the relevant features. Furthermore, during the training of classification models using the reduced feature set, a less computational iterations per second (it/s) were observed. These experimental results clear the concepts about the impact of feature selection techniques, that it not only reduces the size feature space, but it also improves performance of ML models also in various aspects.

\begin{table}[]
\centering
\begin{tabular}{l|c|c|c|c}
\hline
\textbf{Model}                           & \textbf{Accuracy} & \textbf{Balanced Accuracy} & \textbf{ROC AUC} & \textbf{F1-Score} \\ \hline
SVC                             & 0.74     & 0.75              & 0.74    & 0.74     \\ \hline
Nearest Centroid                & 0.74     & 0.75              & 0.74    & 0.74     \\ \hline
Logistic Regression             & 0.73     & 0.74              & 0.73    & 0.73     \\ \hline
SGD Classifier                  & 0.73     & 0.73              & 0.73    & 0.73     \\ \hline
Linear SVC                      & 0.72     & 0.73              & 0.72    & 0.73     \\ \hline
Linear Discriminant Analysis    & 0.72     & 0.73              & 0.72    & 0.73     \\ \hline
Ridge Classifier CV             & 0.72     & 0.73              & 0.72    & 0.73     \\ \hline
Ridge Classifier                & 0.72     & 0.73              & 0.72    & 0.73     \\ \hline
Quadratic Discriminant Analysis & 0.72     & 0.73              & 0.72    & 0.72     \\ \hline
Calibrated Classifier CV        & 0.72     & 0.73              & 0.72    & 0.72     \\ \hline
Label Spreading                 & 0.71     & 0.71              & 0.71    & 0.71     \\ \hline
Bagging Classifier              & 0.70     & 0.70              & 0.70    & 0.70     \\ \hline
AdaBoost Classifier             & 0.70     & 0.71              & 0.70    & 0.71     \\ \hline
Label Propagation               & 0.70     & 0.70              & 0.70    & 0.70     \\ \hline
Bernoulli NB                    & 0.70     & 0.70              & 0.70    & 0.70     \\ \hline
Nu SVC                          & 0.70     & 0.70              & 0.70    & 0.70     \\ \hline
LGBM Classifier                 & 0.70     & 0.70              & 0.70    & 0.70     \\ \hline
Extra Trees Classifier          & 0.69     & 0.70              & 0.69    & 0.69     \\ \hline
XGB Classifier                  & 0.69     & 0.70              & 0.69    & 0.69     \\ \hline
Random Forest Classifier        & 0.69     & 0.70              & 0.69    & 0.69     \\ \hline
Guassian NB                     & 0.69     & 0.69              & 0.69    & 0.69     \\ \hline
Decision Tree Classifier        & 0.68     & 0.69              & 0.68    & 0.69     \\ \hline
Etra Tree Classifier            & 0.68     & 0.69              & 0.68    & 0.69     \\ \hline
KNeighbors Classifier           & 0.67     & 0.68              & 0.67    & 0.68     \\ \hline
Perceptron                      & 0.64     & 0.64              & 0.64    & 0.64     \\ \hline
Passive Agressive Classifier    & 0.63     & 0.63              & 0.63    & 0.63     \\ \hline
Dummy Classifier                & 0.50     & 0.50              & 0.50    & 0.50     \\ \hline
\end{tabular}
\caption{Classification results of various ML models for CVD dataset using reduced feature set}
\label{tab:CVD_redufeat}
\end{table}

\begin{table}[]
\centering
\begin{tabular}{l|c|c|c|c}
\hline
\textbf{Model}                          & \textbf{Accuracy} & \textbf{Balanced Accuracy} & \textbf{ROC AUC} & \textbf{F1-Score} \\ \hline
Perceptron                     & 0.71     & 0.72              & 0.72    & 0.71     \\ \hline
AdaBoost Classifier            & 0.71     & 0.71              & 0.71    & 0.71     \\ \hline
SGD Classifier                 & 0.69     & 0.69              & 0.69    & 0.69     \\ \hline
Logistic Regression            & 0.69     & 0.69              & 0.69    & 0.69     \\ \hline
Bernoulli NB                   & 0.68     & 0.69              & 0.69    & 0.68     \\ \hline
Linear Discriminant Analysis   & 0.68     & 0.68              & 0.68    & 0.68     \\ \hline
Ridge Classifier CV            & 0.68     & 0.68              & 0.68    & 0.68     \\ \hline
Ridge Classifier               & 0.68     & 0.68              & 0.68    & 0.68     \\ \hline
Linear SVC                     & 0.68     & 0.68              & 0.68    & 0.68     \\ \hline
Calibrated Classifier CV       & 0.68     & 0.68              & 0.68    & 0.68     \\ \hline
Gaussian NB                    & 0.66     & 0.68              & 0.68    & 0.65     \\ \hline
SVC                            & 0.67     & 0.67              & 0.67    & 0.67     \\ \hline
Nearest Centroid               & 0.66     & 0.67              & 0.67    & 0.66     \\ \hline
KNeighbors Classifier          & 0.65     & 0.66              & 0.66    & 0.65     \\ \hline
Bagging Classifier             & 0.63     & 0.64              & 0.64    & 0.63     \\ \hline
Quadrant Discriminant Analysis & 0.62     & 0.64              & 0.64    & 0.58     \\ \hline
Decision Tree Classifier       & 0.62     & 0.62              & 0.62    & 0.61     \\ \hline
Extra Tree Classifier          & 0.62     & 0.62              & 0.62    & 0.62     \\ \hline
Random Forest Classifier       & 0.62     & 0.62              & 0.62    & 0.62     \\ \hline
Label Spreading                & 0.59     & 0.59              & 0.59    & 0.59     \\ \hline
Label Propagation              & 0.58     & 0.58              & 0.58    & 0.58     \\ \hline
LGBM Classifier                & 0.57     & 0.58              & 0.58    & 0.57     \\ \hline
Nu SVC                         & 0.57     & 0.58              & 0.58    & 0.57     \\ \hline
XGB Classifier                 & 0.57     & 0.57              & 0.57    & 0.57     \\ \hline
Passive Agressive Classifier   & 0.57     & 0.56              & 0.56    & 0.57     \\ \hline
Dummy Classifier               & 0.55     & 0.56              & 0.56    & 0.55     \\ \hline
Extra Tree Classifier          & 0.51     & 0.51              & 0.51    & 0.51     \\ \hline
\end{tabular}
\caption{Classification results of various ML models for framingham dataset using reduced feature set}
\label{tab:Fram_redufeat}
\end{table}

\section{Conclusion and Future Works}
Heart disease is the most fatal disease which is rapidly increasing and became one of the causes of death around the world. The damage caused by this disease can be reduced significantly, if adequate treatment procedures are applied at the early stages. This paper studies the prediction of heart disease and the selection of the important features. The main goal of this research study is to observe the impact of feature selection techniques on the performance of ML models. This analysis was performed for CVD and Framingham heart disease datasets which are available online. In this research, first, we performed a data pre-processing step in which data transformation, cleansing and balancing were involved. Secondly, we used a filter-based feature selection technique namely the ANOVA-F test to identify the most important features from the datasets for an effective heart disease prediction. Using the ANOVA-F test most relevant features with outcomes from both datasets were identified using the individual feature scores. We observed that features like age, hypertension, glucose, previous heart disease, and blood pressure were found to reflect the most important risk factors for heart disease except the traditional factors using both datasets. Furthermore, the classification experiments were performed with full as well as the reduced feature sets to analyze the effect of selected features on the prediction accuracy of various ML prediction models. Using the full feature set the highest accuracy achieved was 0.73 for CVD and 0.66 for the Framingham heart disease dataset. After using the reduced feature set the accuracy increased to 0.75 and 0.71 for both datasets. The analysis showed that even after limiting the number of features, ML models showed better performance as compared to the models using a full feature set. The experimental results reveal that by employing a feature selection technique, we may accurately classify the heart disease even with a small number of features and less time.  We can conclude that using the feature selection only the most important features related to heart disease are selected which reduces the computational complexities and improve the accuracy of prediction models. In the intended future work, we will try to work on enhancing the prediction accuracy by using a vast combination of ML and deep learning models~\cite{das2021estimating} to obtain the best feasible model for the heart disease diagnosis. We will benchmark our analysis on additional datasets as a part of our future work. We will also try to use more than one feature selection technique to obtain more feasible feature subsets which are more direct with medical studies. 

\label{sec:conclusion}

\section*{Acknowledgments}
This research has received funding from the European Union's Horizon 2020 research and innovation programme under the Marie Skłodowska-Curie grant agreement No. 801522, by Science Foundation Ireland and co-funded by the European Regional Development Fund through the ADAPT Centre for Digital Content Technology grant number 13/RC/2106\_P2.

\end{document}